\documentclass[10pt,twocolumn,letterpaper]{article}
\usepackage[accsupp]{axessibility}  

\usepackage[pagenumbers]{iccv} 
\definecolor{iccvblue}{rgb}{0.21,0.49,0.74}
\usepackage[pagebackref,breaklinks,colorlinks,allcolors=iccvblue]{hyperref}

\usepackage{lipsum}
\usepackage{amsmath,amssymb,amsfonts}
\usepackage{algorithmic}
\usepackage{multirow}
\usepackage{subcaption}
\usepackage{graphicx}
\usepackage{textcomp}
\usepackage{booktabs}
\usepackage{hyperref}

\usepackage{listings}
\usepackage{adjustbox}
\def\BibTeX{{\rm B\kern-.05em{\sc i\kern-.025em b}\kern-.08em
    T\kern-.1667em\lower.7ex\hbox{E}\kern-.125emX}}

\usepackage{pgfplotstable}    
\usepackage{pgfplots}
\usepgfplotslibrary{groupplots,dateplot}
\usetikzlibrary{patterns,shapes.arrows}
\pgfplotsset{compat=newest}
\usepgfplotslibrary{groupplots}

\usepackage[nolist]{acronym}

\DeclareMathOperator*{\argmax}{arg\,max}

\def\paperID{*****} 
\def\confName{ICCV}
\def\confYear{2025}

\title{CTC Transcription Alignment of the Bullinger Letters: Automatic Improvement of Annotation Quality}

\author{Marco Peer$^1$, Anna-Scius Bertrand$^2$ and Andreas Fischer$^1$\\
$^1$University of Fribourg, Switzerland\\
$^2$University of Applied Sciences and Arts Western Switerland, Fribourg, Switzerland\\
{\tt\small marco.peer@unifr.ch, anna.scius-bertrand@hefr.ch, andreas.fischer@unifr.ch}
}

\begin{document}
\newacro{CER}[CER]{Character Error Rate}
\newacro{WER}[WER]{Word Error Rate}
\newacro{HTR}[HTR]{Handwritten Text Recognition}

\definecolor{citecolor}{RGB}{34,139,34}
\definecolor{citecolor2}{HTML}{0071bc}
\definecolor{lightred}{RGB}{241,140,142}
\newcommand{\prel}[1]{{\bf \fontsize{7}{42}\selectfont {\color{lightred!180}~$\downarrow$#1}}}
\newcommand{\mrel}[1]{{\bf \fontsize{7}{42}\selectfont {\color{citecolor!80}~$\uparrow$#1}}}
\newcolumntype{L}[1]{>{\raggedright\arraybackslash}p{#1}}
\newcommand{\crel}[1]{{\bf \fontsize{7}{42}\selectfont {\color{citecolor!80}~$\downarrow$#1}}}
\newcolumntype{L}[1]{>{\raggedright\arraybackslash}p{#1}}

\maketitle
\begin{abstract}
Handwritten text recognition for historical documents remains challenging due to handwriting variability, degraded sources, and limited layout-aware annotations. In this work, we address annotation errors—particularly hyphenation issues—in the Bullinger correspondence, a large 16th-century letter collection. We introduce a self-training method based on a CTC alignment algorithm that matches full transcriptions to text line images using dynamic programming and model output probabilities trained with the CTC loss. Our approach improves performance (e.g., by 1.1 percentage points CER with PyLaia) and increases alignment accuracy. Interestingly, we find that weaker models yield more accurate alignments, enabling an iterative training strategy. We release a new manually corrected subset of 100 pages from the Bullinger dataset, along with our code and benchmarks. Our approach can be applied iteratively to further improve the CER as well as the alignment quality for text recognition pipelines. Code and data are available via \href{https://github.com/andreas-fischer-unifr/nntp}{https://github.com/andreas-fischer-unifr/nntp}.

\end{abstract}

\section{Introduction}

\ac{HTR} remains a challenging task in document analysis, particularly for historical documents, due to the high variability of handwriting styles, the degraded condition of the sources, and the diversity of languages and scripts~\cite{Fischer2020-al}. In addition, the annotation process requires historical expertise to produce accurate transcriptions, which makes it both time-consuming and costly. End-to-end learning approaches based on deep neural networks further increase this issue, as they typically rely on large quantities of annotated training data in the form of text line images paired with their transcriptions~\cite{Jungo2023}. Given the high effort involved in creating such ground truth data, a trade-off often arises between data quality and quantity.

\begin{figure}
\begin{subfigure}{0.97\columnwidth}\centering
        \includegraphics[width=\textwidth]{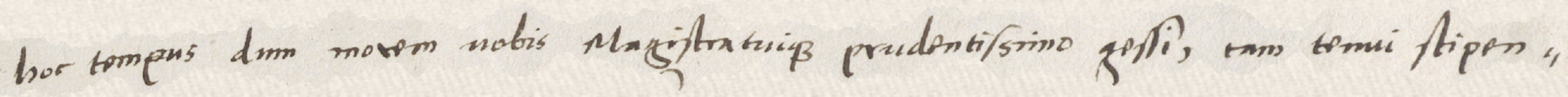}\\
        \includegraphics[width=\textwidth]{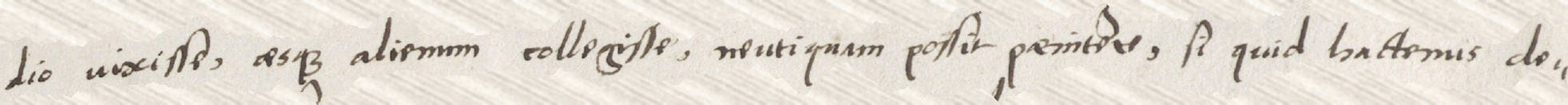}
        \texttt{hoc tempus, [...], tam tenui stipen\textcolor{red}{dio} \\
        vixisse  [...], quid hactenus de\textcolor{red}{cessit,}}
    \caption{}\label{fig:bullinger-example}
\end{subfigure}\\[0.2em]

\begin{subfigure}{0.95\columnwidth}
        \centering
    \includegraphics[width=0.98\columnwidth]{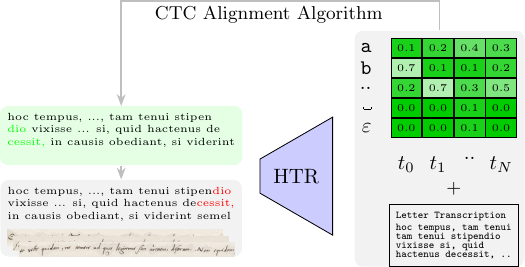}
    \caption{}    \label{fig:abstract}
    \end{subfigure}
    \caption{Bullinger dataset. (a) Examples of alignment errors (\textcolor{red}{red}) in the Bullinger dataset. (ID: \texttt{6\_00\_r1l\{19,20\}}) (b) Overview of our method. We use our CTC alignment algorithm and propose a self-training scheme to improve \ac{HTR} performance to remove errors caused by hyphenations.}
\end{figure}
 
 A key difficulty in the context of acquiring ground truth for training lies in the differing requirements of historical scholarship and computational processing: historians, on the one hand, lean towards regularized versions of texts where abbreviations are resolved and special characters are normalized to suit the reading habits of the contemporary audience, whereas \ac{HTR} systems rely on character-accuracte transcriptions with respect to the visual representation \cite{Mayr2025}. In this work, we address one major challenge of this process through a case study — namely the hyphenation problem due to an automated alignment — on the Bullinger correspondence~\cite{bullinger_dataset}, a large and linguistically diverse collection of 16th-century letters from Switzerland, for which experts have produced accurate transcriptions of the letters of Heinrich Bullinger, an influential Swiss Reformer and theologian, and his correspondences.  However, the current transcriptions are provided only letter-wise without any layout information to enable \ac{HTR} training (transcriptions done by historians for the Bullinger collection are publicly available via GitHub\footnote{\url{https://github.com/bullinger-digital/bullinger-korpus-tei}}). Therefore, to create training data for \ac{HTR}, the Bullinger dataset contributed by Scius-Bertrand et al.\cite{bullinger_dataset} applies an automatic alignment via Transkribus\cite{Muehlberger2019} between these letter-level transcriptions and extracted text line images. While this process enables the creation of a large dataset — over 100,000 line images - it also introduces systematic errors, mainly caused by hyphenations at the beginning and end of lines. Although multiple other errors are present in the ground truth, including misaligned parts of hyphenated words, abbreviations that are written out in full and therefore differ from the image, incorrect capitalization, and mismatched punctuation marks, Jungo et al.\cite{Jungo2023} identified hyphenation as the most common inconsistency within the data collection. It does not only affect the quality of the training data but also the reliability of evaluation results when using the Bullinger dataset for \ac{HTR} research. An example of a wrong hyphenation in the training set is shown in Figure\ref{fig:bullinger-example}. 
 
To tackle the misalignment, improve the annotation quality and study its influence on the Bullinger collection, we introduce a self-training procedure based on a Connectionist Temporal Classification (CTC)~\cite{Graves2006} alignment algorithm in our work, as shown in Figure~\ref{fig:abstract}. It uses dynamic programming to find an optimal sequence with respect to the output probabilities of our \ac{HTR} models trained with the CTC loss function. Furthermore, it takes the full letter transcription (the ground truth - provided by experts, without any layout information), and outputs the aligned text in terms of lines. We align the training set of the Bullinger dataset~\cite{bullinger_dataset} and finetune the models to, as we show in our results, improve the \ac{HTR} - in case of the PyLaia framework by 1.1 percentage points on the CER -- as well as the alignment accuracy. Secondly, we inspect our algorithm and find that the alignment accuracy is even higher when using a weakly trained model. This process can be efficiently repeated iteratively to further enhance the text recognition. To examine our methodology, results are evaluated on a manually created and corrected dataset consisting of 100 pages in total to ensure precise transcription.

To summarize, our contributions are as follows:
\begin{itemize}
    \item We provide a new, manually annotated and corrected dataset of the Bullinger collection,
    \item We introduce a CTC alignment algorithm for text alignment to insert line breaks to match a given layout when ground truth transcriptions are available,
    \item And we propose a self-training strategy to iteratively improve the \ac{HTR} performance when using the alignment method. 
\end{itemize}

The code for the alignment as well as the data to benchmark our algorithm will be publicly available. The remainder of the paper is structured as follows: Section~\ref{sec:related_work} describes related work in the domain of alignment and self-training approaches. Section~\ref{sec:methodology} explains our methodology, the evaluation protocol is reported in Section~\ref{sec:evaluation}, followed by the results in Section~\ref{sec:results}. We conclude our paper in Section~\ref{sec:conclusion}.

\section{Related Work}\label{sec:related_work}
This section briefly describes related work in the general domain of computer vision, followed by historical document analysis.

\subsection{Alignment in computer vision}

Image-text alignment in computer vision refers to the process of establishing meaningful correspondences between visual data and textual descriptions, e.g., in the context of sign language recognition, where sequences of video frames are aligned with gloss-level transcriptions~\cite{min2021visual, zheng2023cvt, kindiroglu2023aligning}, aligning videos with textual descriptions~\cite{shu2022see, wang2023unified, yang2021taco}, or 3D scenes with textual descriptions~\cite{zhu20233d}. The alignment of the visual and the textual modalities serves for automatic transcription, automatic captioning, and cross-modal retrieval, e.g., retrieving a video based on a text or vice versa. Alignment results can also be used for self-learning, using the aligned pseudo-labels for supervised fine-tuning.

\subsection{Alignment for historical manuscripts}

For historical manuscripts, Sudholt and Fink proposed a word-level alignment between images and text with the PHOCNet~\cite{sudholt2017phoc}, a CNN-based network that estimates a pyramidal histogram of characters~\cite{almazan14phoc} representation of text labels. It allows to retrieve keyword images from the (word-segmented) manuscripts using both text queries and image queries.

Most alignment methods in the literature operate at the line level, aiming at simplifying the navigation in digital libraries (switching easily between the transcription and the original handwriting images) and, more commonly, with the goal of leveraging existing human transcriptions of ancient documents to extract line-level pseudo-labels for training \ac{HTR} systems. The methods differ in the optical models they use and the algorithms considered for alignment, which typically include some form of dynamic programming.

Early work includes the alignment method proposed by A. Fischer et al.~\cite{fischer2011transcription} for historical Latin manuscripts, which is based on character hidden Markov models (HMMs) and Viterbi decoding for alignment. Spelling variants are included in the models and it is demonstrated that the alignment task can be solved with high accuracy even for weakly trained models using only few annotated learning samples.

More recently, Ezra et al.~\cite{ezra2020transcription} proposed an alignment method for fragments of the Dead Sea Scrolls. After line segmentation, a CNN-RNN based \ac{HTR} system is used to produce an automatic recognition result, which is then aligned with the transcription using string edit distance. In the work of P. Torras et al.~\cite{torras2021transcription}, a Seq2Seq model is used to align historical handwritten ciphers based on attention mechanisms. The approach of B. Madi et al.~\cite{madi2022textline} targets the task of image-image alignment between Arabic manuscripts that contain similar texts. YOLO object detection is considered for spotting subwords, subword similarity is established by means of siamese CNN, and alignment is achieved using a longest common subsequence algorithm. YOLO object detection is also considered in the work of A. Scius-Bertrand et al.~\cite{scius2021transcription} for aligning columns of ancient Vietnamese manuscripts with their transcription, utilizing clustering algorithms for the alignment. An interesting aspect of their work is that neither page segmentation nor human-annotated learning samples are needed for aligning the logographic characters.

In this paper, we introduce a line-level alignment method that is based on a deep CNN-RNN with CTC loss. Instead of relying on the recognition result as basis for the alignment, as in~\cite{ezra2020transcription}, we leverage the character CTC posterior probabilities, similar to the character HMM likelihoods in~\cite{fischer2011transcription}, and perform a Viterbi-like dynamic programming to find the optimal correspondence between the character probabilities and the transcription. The CTC character probabilities can be seen as a preliminary character-level image-text alignment performed by the CNN-RNN, which is then used for aligning the transcription of entire letters, consisting of multiple scanned pages.

\section{Methodology}\label{sec:methodology}
In this section, we present our methodology, starting with the manual annotation and correction of two subsets of the Bullinger collection\footnote{\url{https://www.bullinger-digital.ch/index.html}}. Then, we present the CTC alignment algorithm proposed in our work, followed by the approaches used for \ac{HTR}. Lastly, the self-training procedure is explained.

\subsection{Bullinger Dataset}

The Bullinger dataset~\cite{bullinger_dataset} is a historical manuscript collection consisting of 3,622 letters written by 306 authors, drawn from the larger correspondence of Swiss reformer Heinrich Bullinger (1504-–1575). The dataset reflects the linguistic diversity of the period, with letters primarily written in Latin and premodern German, alongside passages in Greek, Italian, French, and Hebrew. Transcriptions are not strictly character-accurate, as abbreviations are often expanded, and the handwriting - particularly that of Bullinger himself — can be highly variable and difficult to read, posing challenges for both human annotators and \ac{HTR} systems. Additionally, due to the automated alignment, done with the Transkribus platform~\cite{Muehlberger2019}, errors arise at the beginning and end of text lines, particularly in the presence of word breaks.

Therefore, we create a manually annotated and corrected dataset of 100 pages, consisting of 2,388 text lines, split into two subsets. The first subset contains transcriptions made by historians, reflecting how they interpret the text. These transcriptions are written as continuous paragraphs rather than line by line; hyphenated words at line breaks in the images are not preserved. Written abbreviations are expanded, and punctuation is added. This subset also represents the workflow of the annotations provided in the Bullinger dataset~\cite{bullinger_dataset}.

The second subset contains what we call diplomatic transcriptions, meaning they closely reflect the original line-by-line content of the images. Hyphenated words are preserved, abbreviations are kept as written, and no modern punctuation is added. The dataset statistics are shown in Table~\ref{tab:our_dataset}.

\begin{table}[] 
\centering \small
\begin{tabular}{p{3cm}ccc}
\toprule
 ~ & Letters & Pages & Lines \\
\midrule
Subset 1 & 20 & 50 & 902\\
Subset 2 & 49 & 50  & 1486\\ \midrule
Total & 69 & 100 & 2388 \\
\bottomrule
\end{tabular}
\caption{Statistics of the manually corrected dataset.}\label{tab:our_dataset}
\end{table}

\subsection{Handwritten Text Recognition}

We use three different models for \ac{HTR}: HTRFlor~\cite{deSousaNeto2020} (820k parameters), PyLaia~\cite{PyLaia} (6.4M), and the method by Retsinas et al.~\cite{Retsinas2022} (7.4M). While the use of HTRFlor and PyLaia follows the approach of the original paper~\cite{bullinger_dataset}, the method by Retsinas et al. presents a trade-off between training speed and number of parameters. It retains image aspect ratio, replaces max-pooling with column-wise concatenation to reduce parameters, and introduces a parallel shortcut branch with a 1D convolution and CTC~\cite{Graves2006} loss to guide the CNN backbone toward learning more discriminative features. 
The general structure is a CNN-based backbone, followed by either a block of BiLSTM~\cite{Hochreiter1997} (Retsinas et al., PyLaia) or BGRU~\cite{Cho2014} (HTRFlor) layers. For our work, we follow the default architectures and training strategies proposed by the authors. 
The architectures of our models are shown in the supplementary material (Figure~\ref{fig:architecture-comparison}).

\subsection{CTC Alignment}

The proposed image-text alignment method is inspired by the NNTP token passing algorithm introduced in~\cite{Fischer2012} for \ac{HTR} using very large vocabularies. Our method is designed for \ac{HTR} systems that are based on the CTC loss function~\cite{Graves2006}. Such systems provide estimated character posterior probabilities $P(c_i|t_j)$ for all characters $c_i \in \mathcal{A}$ of the alphabet $\mathcal{A}$, including the special $\varepsilon$ character indicating ``no character'', for all time steps $t_j$ when processing the convolutional features of text line images using recurrent neural networks (LSTM or GRU, respectively). While the probabilities are estimated bi-directionally from left-to-right and from right-to-left, the time axis corresponds to the natural reading order of the handwriting.

\paragraph{Problem statement}
The alignment algorithm receives two inputs: the machine-readable transcription of the letter, and the character posterior probabilities for all text line images of the letter. The goal is to insert newline characters at the correct position within the transcription, such that it is aligned with the text line images, resulting in labeled samples for training \ac{HTR} systems. For each aligned text line, the average character probability, as well as the average probability over the first six and the last six characters is returned as measures of confidence.

Note that in this problem statement, the transcription itself is not changed, i.e. it is not possible to skip words, change or add characters, swap parts of the text, etc. However, if there are parts of the image that are missing in the transcription, the alignment can insert multiple newline characters in order to create a gap in the aligned transcription. 

\paragraph{Alignment process}

\begin{figure}
    \centering
    \includegraphics[width=0.95\columnwidth]{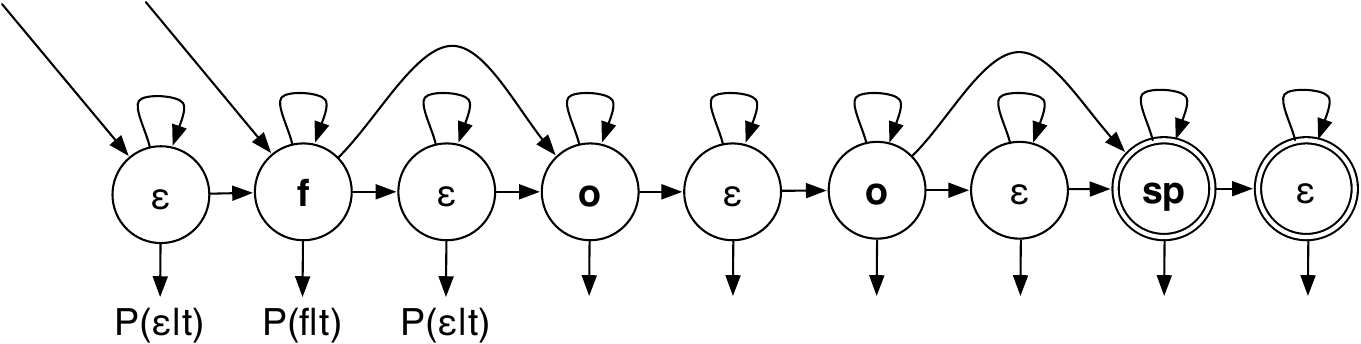}
    \caption{FSA representing the transcription ``foo '' after $\varepsilon$ extension (`sp' is the whitespace character). Each character state emits posterior probabilities with respect to the time $t$. There are two initial states (diagonal arrows) and two final states (double circles).}
    \label{fig:fsa}
\end{figure}

The alignment proceeds in three steps. In the first step, both the letter transcription and the character posterior probabilities are preprocessed. After discarding all newline characters from the letter transcription, following the principle of CTC, the transcription is extended to a linear, Finite State Automaton (FSA) with skip connections that also include the $\varepsilon$ character, as illustrated in Figure~\ref{fig:fsa}. Note that there is no direct connection between two identical characters. At least one $\varepsilon$ needs to be visited between them, such that a character transition can be detected. Regarding the character probabilities, they are concatenated over all lines and pages of a handwritten letter, in order to form a single probability sequence for the entire letter. Afterwards, we compress the sequence by means of a threshold $\theta_\varepsilon$, such that successive time steps with $P(\varepsilon|t_j) > \theta_\varepsilon$ are compressed into a single time step with probability
\[
P(\varepsilon|t_{ab}) = \prod_{j=a}^b P(\varepsilon|t_j)\ ,
\]
and similar for all other characters of the alphabet. This $\varepsilon$ compression is inspired by the observation that for CTC, the $\varepsilon$ character is almost always ``active'' with nearly 100\% probability, interrupted only sporadically by time steps, in which standard characters have a probability significantly larger than 0\%.

\begin{figure}
    \centering
    \includegraphics[width=0.95\columnwidth]{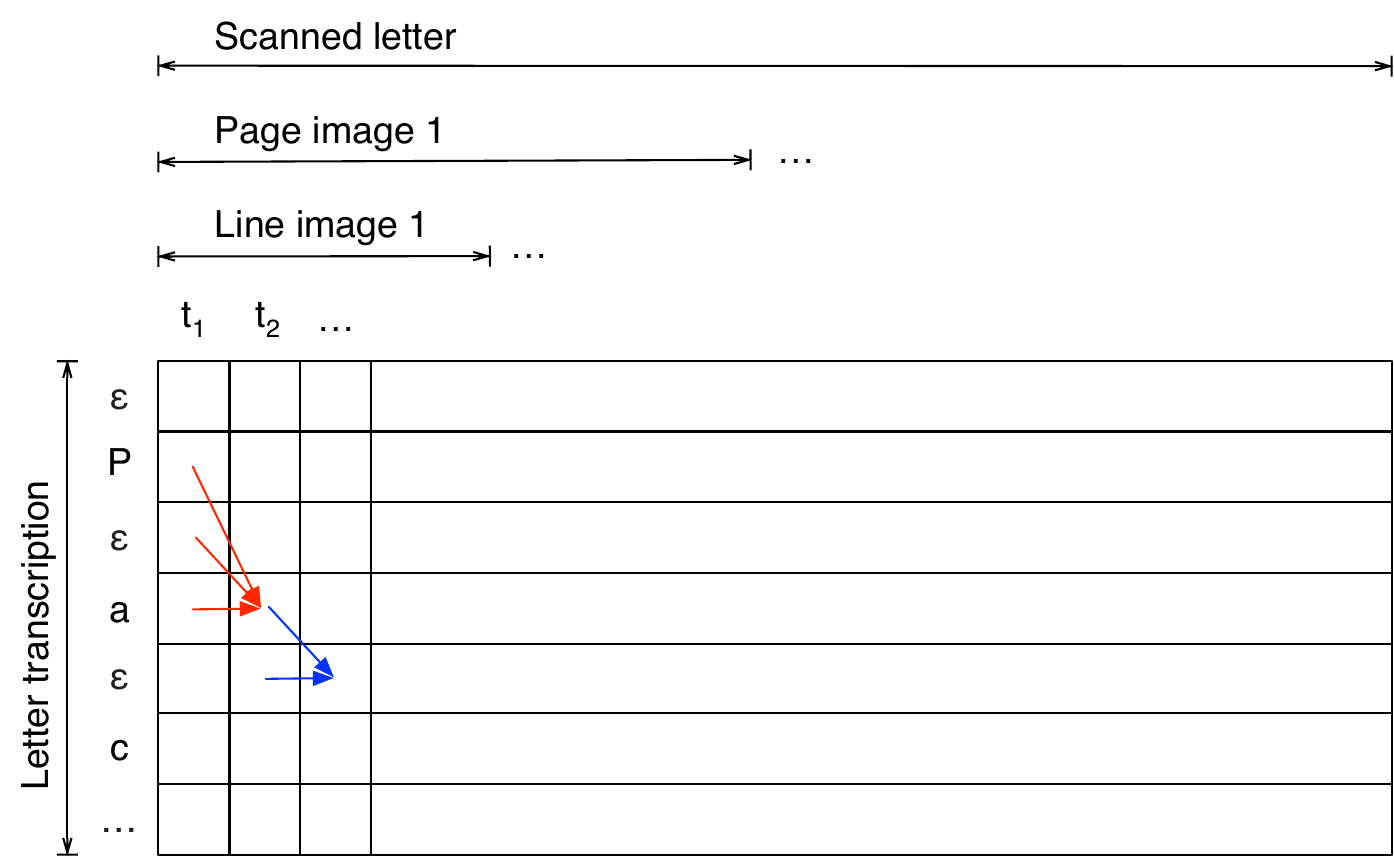}
    \caption{Schema of the dynamic programming algorithm proposed for letter transcription alignment based on CTC character probabilities. Red arrows illustrate the maximum probability step for standard characters, blue arrows for the $\varepsilon$ character.}
    \label{fig:dp}
\end{figure}

In the second, main step, dynamic programming is used to find the optimal character sequence $c^*=c^*_1,\ldots,c^*_T$ with $c^*_j \in \mathcal{A}$, and $T$ the length of the ($\varepsilon$-compressed) letter probability sequence, such that
\[
c^* = \argmax_{c_1,\ldots,c_T \in \mathcal{C}_{FSA}} \prod_{j=1}^T P(c_j|t_j)\ ,
\]
where $\mathcal{C}_{FSA}$ contains all valid character sequences according to the FSA of the text transcription (see Figure~\ref{fig:fsa}). A schema of the dynamic programming procedure is illustrated in Figure~\ref{fig:dp}, highlighting the maximum probability step for standard characters (3 possible predecessors) and the $\varepsilon$ character (2 possible predecessors) according to the FSA. To avoid numerical problems with small probabilities, the algorithm considers sums of logarithms. For increased memory efficiency, token passing is applied, i.e. only one column is kept in memory in form of the FSA and starting from the initial states, tokens are passed to valid successors in the automaton at each time step. At time $T$, the tokens in the final states hold the optimal solution.

In the third and final step, newline characters are inserted in the transcription according to the line break positions of the text line images in the optimal character sequence $c^*$. Furthermore, the two confidence measures
\begin{align*}
\gamma & =\frac{1}{e-s}\sum_{j=s}^{e}P(c^*_j|t_j)\ ,\\
\gamma_{6} & =\frac{1}{6}\sum_{j=s}^{s+6}P(c^*_j|t_j)+\frac{1}{6}\sum_{j=e-6}^{e}P(c^*_j|t_j)\ ,
\end{align*}
are calculated, where $t_s$ is the starting time and $t_e$ is the ending time of the aligned text line. They reflect the quality of the aligned learning samples with respect to the average character probability, focusing on the beginning and the end of the line in the case of $\gamma_6$. For small text lines with less than 12 characters, $\gamma_6$ is set to zero.

\subsection{Self-Training}

We can now address alignment errors by implementing the self-training pipeline. The procedure begins by extracting the probabilities for $P(c_i|t_j), \ c_i \in \mathcal{C}_{FSA}$ for each timestep from the three different \ac{HTR} models. These probabilities are combined with expert-provided letter transcriptions that lack layout information. Using these inputs, the CTC alignment algorithm performs sequence alignment by inserting line breaks to the corresponding characters in the transcription.


Following alignment, we generate training subsets by applying filtering based on $\gamma_6$, the confidence of the alignment for the first and last six characters. For example, we retain only samples that meet a minimum average character confidence for $\gamma_6$. The filtered subsets of the Bullinger's training and validation set are subsequently used to finetune the \ac{HTR} models. This adaptation is aimed to improve the \ac{HTR} performance, in particular when alignment issues, such as hyphenation, are occuring. We show that the procedure can be applied iteratively - improved model predictions are used to generate refined alignments, which in turn support further finetuning.
\section{Evaluation}\label{sec:evaluation}
\subsection{Setup}
\paragraph{Data} We use the original Bullinger dataset~\cite{bullinger_dataset} and remove the letters which occur in our contributed test set, resulting in 108.7k lines for training (instead of 109.6k). The alphabet to generate $\mathcal{C}_{FSA}$ consists of the 79 symbols of the training set (including the $\varepsilon$ of the CTC). 

\paragraph{Training and Finetuning} We use the defaults parameters for training and model architecture of the three frameworks. For finetuning, we decrease the number of epochs to 10\% of the initial training phase and lower the learning rate by $0.01$. In both phases, a learning rate scheduler is used, and all of our models converge. 

\paragraph{CTC Alignment} The threshold for the $\theta_\varepsilon$ is set to 0.99 in our experiments. If not stated otherwise, we use the PyLaia model for \ac{HTR} and finetune it on the alignments filtered with $\gamma_6 > 0.5$.

\subsection{Metrics}
In the following, we briefly describe the metrics used for evaluation.
\paragraph*{Line-level Accuracy} To quantitatively evaluate the quality of predicted text lines against the ground truth in our handwritten text recognition system, we adopt a line-level accuracy metric. For each letter \(i\) (note that a letter usually consists of more than one page), the transcription is segmented into lines by our algorithm and compared to the corresponding ground truth lines in the same order. Let the ground truth lines be $
G_i = \big(g_i^{(1)}, g_i^{(2)}, \ldots, g_i^{(L_i)}\big)$, where \(L_i\) is the number of ground truth lines, and the predicted lines be $P_i = \big(p_i^{(1)}, p_i^{(2)}, \ldots, p_i^{(M_i)}\big)$, where \(M_i\) is the number of predicted lines. We define a per-line match indicator
\begin{equation}
    \delta_i^{(j)} = 
\begin{cases}
1 & \text{if } p_i^{(j)} = g_i^{(j)} \\
0 & \text{otherwise}
\end{cases}
\quad \text{for } j = 1, \ldots, \min(L_i, M_i).
\end{equation}

The number of correctly matched lines for sample \(i\) is then
\begin{equation}
m_i = \sum_{j=1}^{\min(L_i, M_i)} \delta_i^{(j)}.
\end{equation}
Finally, the overall line-level accuracy across the dataset of \(N\) samples is computed as the total number of correctly matched lines normalized by the total number of ground truth lines.
\paragraph*{\ac{HTR}} For evaluating the performance of the HTR models, we provide \ac{CER} and \ac{WER} metrics. Additionally, we report the \ac{CER} metric for the first and last $n$ characters of a line, denoted as \ac{CER}$_n$, to specifically evaluate the influence of the text alignment on the \ac{HTR} performance.

\section{Results}\label{sec:results}
In this part, we present the main results of the proposed methodology.

\begin{table}[htbp] 
\centering \small
\begin{tabular}{p{3cm}cccc}
\toprule
\multirow{2}{*}{} & \multicolumn{2}{c}{Frequent} & \multicolumn{2}{c}{Nonfrequent} \\
                 & \ac{CER} & \ac{WER} & \ac{CER} & \ac{WER} \\ 
\midrule

HTRFlor~\cite{deSousaNeto2020} & 11.8 & 43.7 & 12.8 & 46.4 \\
PyLaia~\cite{PyLaia} & \textbf{\phantom{0}8.9} & \textbf{31.5} & \textbf{\phantom{0}9.9} & \textbf{35.0} \\ 
Retsinas et al.~\cite{Retsinas2022} & \phantom{0}9.0 & 29.6 & 10.3 & 33.7 \\ \midrule
Baseline~\cite{bullinger_dataset} & \phantom{0}8.4 & 29.6 & \phantom{0}9.9 & 34.4 \\
\bottomrule
\end{tabular}
\caption{\ac{HTR} performance on the two writer subsets of the Bullinger dataset~\cite{bullinger_dataset}.}\label{tab:bullinger-baseline}
\end{table}

\subsection{Baseline Results for Bullinger}
First, we evaluate our three \ac{HTR} models on the Bullinger dataset~\cite{bullinger_dataset} and compare it with the results provided in the original work, shown in Table~\ref{tab:bullinger-baseline}. The worst model is HTRFlor (with the fewest number of parameters), followed by Retsinas et al. and PyLaia, that performs best with 8.9~\% / 9.9~\% CER on the frequent/nonfrequent subset of the dataset. Scius-Bertrand et al.~\cite{bullinger_dataset} also provide the binarized version of the line images to the input (resulting in a four channel image), that might explain the small gap in performance.

\begin{table}[] 
\centering \small
\begin{tabular}{p{3cm}cccc}
\toprule
\multirow{2}{*}{} & \multicolumn{2}{c}{Subset 1} & \multicolumn{2}{c}{Subset 2} \\
                 & \ac{CER} & \ac{WER} & \ac{CER} & \ac{WER} \\ 
\midrule
HTRFlor~\cite{deSousaNeto2020} & 15.7 & 52.4 & 22.8 & 64.9 \\
PyLaia~\cite{PyLaia} & \textbf{\phantom{0}7.6} & \textbf{28.9} & \textbf{18.6} & \textbf{54.0} \\
Retsinas et al.~\cite{Retsinas2022} & 13.0 & {40.1} & 21.7 & {58.6} \\
\bottomrule
\end{tabular}
\caption{\ac{HTR} performance on the 100 manually corrected pages.}\label{tab:baseline_results}
\end{table}
Next, we evaluate the same models as in Table~\ref{tab:bullinger-baseline} on our manually corrected dataset, which consists of two subsets of 50 pages each. The results are presented in Table~\ref{tab:baseline_results}. PyLaia outperforms the other two approaches by 5.4/8.1\% and 3.1/4.2\% CER on the two subsets, respectively. However, the performance on Subset 2 is consistently worse across all approaches, which may be due to differences in annotation style. The models were trained on Annotation Set 1, where abbreviations were expanded and punctuation was added. As a result, the models tend to reproduce this normalized format rather than the original transcription as it appears in Subset 2.

\subsection{Text Alignment}

We evaluate the toking passing algorithm by reporting the line-level accuracy on both subsets in Table~\ref{tab:accuracy_results}. All models achieve accuracies above 80 and 70\%, resp., on both of our subsets. Similarly to the \ac{HTR} performance, the results are worse for Subset 2. Interestingly, the smallest model, HTRFlor, outperforms PyLaia, the best model in terms of \ac{CER}, on Subset 2. We will evaluate the influence of the model size in our ablation study. 

\begin{table}[] 
\centering \small
\begin{tabular}{lcc}
\toprule
~ & Subset 1 & Subset 2 \\
\midrule
HTRFlor~\cite{deSousaNeto2020} & \textbf{86.1} & \textbf{79.7}\\ 
PyLaia~\cite{PyLaia} & \textbf{86.1} & 77.3 \\
Retsinas et al.~\cite{Retsinas2022} & 81.5 & 73.0 \\
\bottomrule
\end{tabular}
\caption{Line-level alignment accuracy.}\label{tab:accuracy_results}
\end{table}

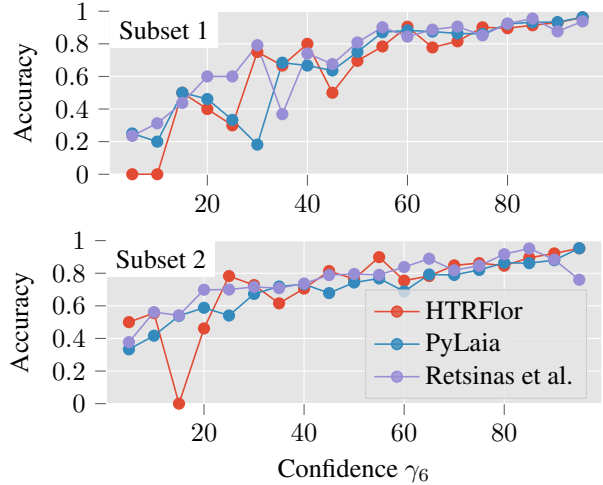
\begin{figure}
    \centering
\begin{tikzpicture}

\definecolor{chocolate2267451}{RGB}{226,74,51}
\definecolor{dimgray85}{RGB}{85,85,85}
\definecolor{gainsboro229}{RGB}{229,229,229}
\definecolor{lightgray204}{RGB}{204,204,204}
\definecolor{mediumpurple152142213}{RGB}{152,142,213}
\definecolor{steelblue52138189}{RGB}{52,138,189}

\begin{axis}[
axis background/.style={fill=gainsboro229},
axis line style={white},
height=0.45\columnwidth,
width=0.98\columnwidth,
legend cell align={left},
legend style={
  fill opacity=0.8,
  draw opacity=1,
  text opacity=1,
  at={(0.97,0.03)},
  anchor=south east,
  draw=lightgray204,
  fill=gainsboro229
},
tick align=outside,
tick pos=left,
title={Subset 1},
every axis title/.style={below right,at={(0,1)},fill=white},
x grid style={white},
xmajorgrids,
xmin=0.5, xmax=99.5,
xtick style={color=dimgray85},
y grid style={white},
ylabel={Accuracy},
ymajorgrids,
ymin=0, ymax=1.0,
ytick style={color=dimgray85}
]
\addplot [semithick, chocolate2267451, mark=*, mark size=2, mark options={solid}]
table {%
5 0
10 0
15 0.5
20 0.4
25 0.3
30 0.75
35 0.666666666666667
40 0.8
45 0.5
50 0.695652173913043
55 0.783783783783784
60 0.904761904761905
65 0.777777777777778
70 0.815384615384615
75 0.901098901098901
80 0.895833333333333
85 0.913669064748201
90 0.930635838150289
95 0.961538461538462
};
\addplot [semithick, steelblue52138189, mark=*, mark size=2, mark options={solid}]
table {%
5 0.25
10 0.2
15 0.5
20 0.461538461538462
25 0.333333333333333
30 0.181818181818182
35 0.684210526315789
40 0.666666666666667
45 0.636363636363636
50 0.75
55 0.870967741935484
60 0.882352941176471
65 0.875
70 0.863636363636364
75 0.860759493670886
80 0.923076923076923
85 0.932038834951456
90 0.933333333333333
95 0.963235294117647
};
\addplot [semithick, mediumpurple152142213, mark=*, mark size=2, mark options={solid}]
table {%
5 0.235294117647059
10 0.3125
15 0.4375
20 0.6
25 0.6
30 0.791666666666667
35 0.368421052631579
40 0.740740740740741
45 0.675675675675676
50 0.808510638297872
55 0.901639344262295
60 0.844444444444444
65 0.887096774193548
70 0.905660377358491
75 0.851851851851852
80 0.923913043478261
85 0.954954954954955
90 0.876543209876543
95 0.9375
};
\end{axis}

\end{tikzpicture}
\begin{tikzpicture}

\definecolor{chocolate2267451}{RGB}{226,74,51}
\definecolor{dimgray85}{RGB}{85,85,85}
\definecolor{gainsboro229}{RGB}{229,229,229}
\definecolor{lightgray204}{RGB}{204,204,204}
\definecolor{mediumpurple152142213}{RGB}{152,142,213}
\definecolor{steelblue52138189}{RGB}{52,138,189}

\begin{axis}[
axis background/.style={fill=gainsboro229},
axis line style={white},
height=0.45\columnwidth,
width=0.98\columnwidth,
legend cell align={left},
legend style={
  fill opacity=0.8,
  draw opacity=1,
  text opacity=1,
  at={(0.97,0.03)},
  anchor=south east,
  draw=lightgray204,
  fill=gainsboro229
},
tick align=outside,
tick pos=left,
title={Subset 2},
every axis title/.style={below right,at={(0,1)},fill=white},
x grid style={white},
xlabel={Confidence $\gamma_6$},
xmajorgrids,
xmin=0.5, xmax=99.5,
xtick style={color=dimgray85},
y grid style={white},
ylabel={Accuracy},
ymajorgrids,
ymin=0, ymax=1.0,
ytick style={color=dimgray85}
]
\addplot [semithick, chocolate2267451, mark=*, mark size=2, mark options={solid}]
table {%
5 0.5
10 0.555555555555556
15 0
20 0.461538461538462
25 0.782608695652174
30 0.727272727272727
35 0.615384615384615
40 0.705882352941177
45 0.813333333333333
50 0.764705882352941
55 0.89873417721519
60 0.754716981132076
65 0.783018867924528
70 0.848214285714286
75 0.862068965517241
80 0.846153846153846
85 0.894736842105263
90 0.921428571428571
95 0.952941176470588
};
\addlegendentry{HTRFlor}
\addplot [semithick, steelblue52138189, mark=*, mark size=2, mark options={solid}]
table {%
5 0.333333333333333
10 0.416666666666667
15 0.538461538461538
20 0.588235294117647
25 0.540540540540541
30 0.673076923076923
35 0.71875
40 0.732394366197183
45 0.67816091954023
50 0.743119266055046
55 0.767857142857143
60 0.688679245283019
65 0.792079207920792
70 0.789473684210526
75 0.82
80 0.864077669902913
85 0.862068965517241
90 0.879699248120301
95 0.953488372093023
};
\addlegendentry{PyLaia}
\addplot [semithick, mediumpurple152142213, mark=*, mark size=2, mark options={solid}]
table {%
5 0.377358490566038
10 0.56140350877193
15 0.541666666666667
20 0.698630136986301
25 0.7
30 0.715909090909091
35 0.708333333333333
40 0.736842105263158
45 0.788732394366197
50 0.795454545454545
55 0.790123456790123
60 0.837837837837838
65 0.888888888888889
70 0.818181818181818
75 0.846153846153846
80 0.916666666666667
85 0.951807228915663
90 0.883333333333333
95 0.76
};
\addlegendentry{Retsinas et al.}
\end{axis}

\end{tikzpicture}
    \caption{Confidence values of the first and last six characters vs. line-level accuracy. Higher confidence values correspond to higher line-level accuracies.}
    \label{fig:accuracy-per-confidence}
\end{figure}

In Figure~\ref{fig:accuracy-per-confidence}, we show the line-level accuracy for each approach depending on the confidence of the first/last six characters on our test set. We notice that higher confidence scores also correspond to higher line-level accuracies, with Retsinas for Subset 2 being the only outlier for higher confidence thresholds, explaining the lower accuracy in Table~\ref{tab:accuracy_results}. 
\begin{table}[htbp] 
\centering \small
\begin{tabular}{lccccc}
\toprule
~ & $T_\mathrm{Line}$ & $T^*_\mathrm{L}$ & $T_\mathrm{L}$ & $\varepsilon$ & $d$ in s\\
\midrule
HTRFlor~\cite{deSousaNeto2020} & 256 & 11242 & 6897 & 1.63 & 3.54\\ 
PyLaia~\cite{PyLaia} & 214 & 8915 & 5643 & 1.58 & 2.70\\
Retsinas et al.~\cite{Retsinas2022} & 126 & 5759 & 4840 & 1.19 & 2.28\\
\bottomrule
\end{tabular}
\caption{Average length of the line ($T_\mathrm{Line}$) and the letter probability sequence ($T^*_\mathrm{L}$/$T_\mathrm{L}$ after/before the compression), the achieved compression ratio $\varepsilon$ when applying thresholding with $\theta_\varepsilon = 0.99$ and the average run-time per letter $d$ for different recognition models.}\label{tab:compression}
\end{table}

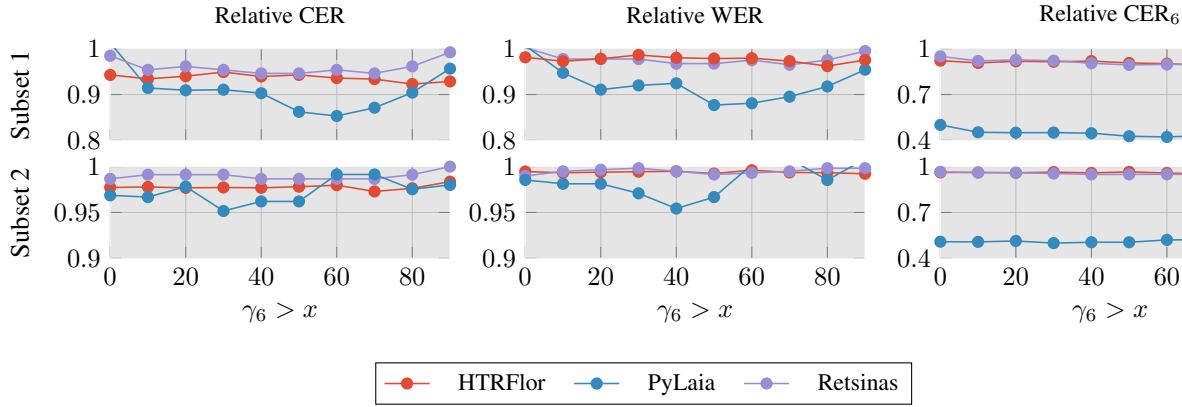
\begin{figure*}[t]
\centering
    \begin{tikzpicture}
\definecolor{chocolate2267451}{RGB}{226,74,51}
\definecolor{dimgray85}{RGB}{85,85,85}
\definecolor{gainsboro229}{RGB}{229,229,229}
\definecolor{lightgray204}{RGB}{204,204,204}
\definecolor{mediumpurple152142213}{RGB}{152,142,213}
\definecolor{steelblue52138189}{RGB}{52,138,189}

    \begin{groupplot}[group style={group size=3 by 2,    vertical sep=0.35cm}, height=2.8cm, width=6.1cm, ymin=0.80, ymax=1,xmin=0,xmax=90,ytick={0.8,0.9,1},
    axis background/.style={fill=gainsboro229},
    axis line style={white},
    xmajorgrids,
    ymajorgrids,
    every axis title/.append style={font=\small},
  legend style={at={(0.5,1)},column sep=0.3cm, 
  anchor=north, legend columns=-1, font=\small}, 
  legend to name=mylegend]
    
        \nextgroupplot[title=Relative CER, xticklabels={}]
            \addplot[semithick, chocolate2267451, mark=*, mark size=2, mark options={solid}] coordinates {
    (0,0.942857) 
    (10,0.934286) 
    (20,0.94) 
    (30,0.949143) 
    (40,0.939429) 
    (50,0.942857) 
    (60,0.936) 
    (70,0.933714) 
    (80,0.922857) 
    (90,0.928571)
};
\addplot[semithick, steelblue52138189, mark=*, mark size=2, mark options={solid}]  coordinates {
    (0,1.012771)
    (10,0.914432)
    (20,0.909323)
    (30,0.910600)
    (40,0.902937)
    (50,0.862237)
    (60,0.853129)
    (70,0.871234)
    (80,0.904215)
    (90,0.956577)
};
\addplot [semithick, mediumpurple152142213, mark=*, mark size=2, mark options={solid}] coordinates {
    (0,0.984615)
    (10,0.953846)
    (20,0.961538)
    (30,0.953846)
    (40,0.946154)
    (50,0.946154)
    (60,0.953846)
    (70,0.946154)
    (80,0.961538)
    (90,0.992308)
};

\nextgroupplot[title=Relative WER, xticklabels={}]
        \addplot[semithick, mediumpurple152142213, mark=*, mark size=2, mark options={solid}] coordinates {
    (0,1.002493)
    (10,0.977557)
    (20,0.977557)
    (30,0.977557)
    (40,0.967581)
    (50,0.967581)
    (60,0.975062)
    (70,0.965087)
    (80,0.975062)
    (90,0.995012)
};
\addplot[semithick, steelblue52138189, mark=*, mark size=2, mark options={solid}]  coordinates {
    (0,1.006865)
    (10,0.947368)
    (20,0.910755)
    (30,0.919908)
    (40,0.924485)
    (50,0.877011)
    (60,0.881007)
    (70,0.89534)
    (80,0.917620)
    (90,0.954233)
};
\addplot[semithick, chocolate2267451, mark=*, mark size=2, mark options={solid}]  coordinates {
    (0,0.981346)
    (10,0.972695)
    (20,0.978102)
    (30,0.986483)
    (40,0.980265)
    (50,0.978643)
    (60,0.979454)
    (70,0.972966)
    (80,0.962152)
    (90,0.975399)
};
\nextgroupplot[ymin=0.4,title=Relative CER$_6$, xticklabels={}, ytick={0.4,0.7,1}]
\addplot[semithick, chocolate2267451, mark=*, mark size=2, mark options={solid}]  coordinates {
    (0,0.922256)
    (10,0.907012)
    (20,0.917683)
    (30,0.914634)
    (40,0.919207)
    (50,0.907012)
    (60,0.902439)
    (70,0.890244)
    (80,0.891768)
    (90,0.878049)
};
\addplot[semithick, steelblue52138189, mark=*, mark size=2, mark options={solid}]  coordinates {
    (0,0.501271)
    (10,0.452598)
    (20,0.450069)
    (30,0.450701)
    (40,0.446909)
    (50,0.426764)
    (60,0.422256)
    (70,0.431262)
    (80,0.447541)
    (90,0.473458)
};
 \addplot[semithick, mediumpurple152142213, mark=*, mark size=2, mark options={solid}] coordinates {
    (0,0.950)
    (10,0.920)
    (20,0.927)
    (30,0.922)
    (40,0.905)
    (50,0.894)
    (60,0.897)
    (70,0.898)
    (80,0.912)
    (90,0.953)
};
\nextgroupplot[ymin=0.9,xlabel=$\gamma_6 > x$, ytick={0.9,0.95,1}]
\addplot[semithick, chocolate2267451, mark=*, mark size=2, mark options={solid}]  coordinates {
    (0,0.977227)
    (10,0.977828)
    (20,0.976790)
    (30,0.977227)
    (40,0.976899)
    (50,0.978046)
    (60,0.979739)
    (70,0.972912)
    (80,0.976298)
    (90,0.983616)
};\addlegendentry{HTRFlor}
    \addplot[semithick, mediumpurple152142213, mark=*, mark size=2, mark options={solid}] coordinates {
    (0,0.986639)
    (10,0.991242)
    (20,0.991242)
    (30,0.991242)
    (40,0.986639)
    (50,0.986639)
    (60,0.986639)
    (70,0.986639)
    (80,0.991242)
    (90,1.000000)
};
\addplot[semithick, steelblue52138189, mark=*, mark size=2, mark options={solid}]  coordinates {
    (0,0.968661)
    (10,0.966762)
    (20,0.978158)
    (30,0.951567)
    (40,0.962013)
    (50,0.962013)
    (60,0.991453)
    (70,0.991453)
    (80,0.975309)
    (90,0.980057)
};

        \nextgroupplot[xlabel=$\gamma_6 > x$,ymin=0.9,ytick={0.9,0.95,1}]
\addplot[semithick, chocolate2267451, mark=*, mark size=2, mark options={solid}]  coordinates {
    (0,0.994675)
    (10,0.993286)
    (20,0.993981)
    (30,0.994444)
    (40,0.994907)
    (50,0.992476)
    (60,0.996064)
    (70,0.993634)
    (80,0.993402)
    (90,0.992245)
};
\addplot[semithick, mediumpurple152142213, mark=*, mark size=2, mark options={solid}] coordinates {
    (0,0.989761)
    (10,0.994875)
    (20,0.996647)
    (30,0.998292)
    (40,0.994875)
    (50,0.991457)
    (60,0.993120)
    (70,0.994875)
    (80,0.998292)
    (90,0.998292)
};
\addplot[semithick, steelblue52138189, mark=*, mark size=2, mark options={solid}]  coordinates {
    (0,0.985447)
    (10,0.981289)
    (20,0.981289)
    (30,0.970894)
    (40,0.954262)
    (50,0.966736)
    (60,1.002079)
    (70,1.008378)
    (80,0.985447)
    (90,1.008316)
};
\nextgroupplot[ymin=0.4,ytick={0.4,0.7,1},xlabel=$\gamma_6 > x$]
\addplot[semithick, chocolate2267451, mark=*, mark size=2, mark options={solid}]  coordinates {
    (0,0.961988)
    (10,0.961988)
    (20,0.959795)
    (30,0.964181)
    (40,0.959064)
    (50,0.965643)
    (60,0.958333)
    (70,0.951023)
    (80,0.948099)
    (90,0.945906)
};\addlegendentry{HTRFlor}
\addplot[semithick, steelblue52138189, mark=*, mark size=2, mark options={solid}]  coordinates {
    (0,0.508152)
    (10,0.507155)
    (20,0.513133)
    (30,0.499184)
    (40,0.504664)
    (50,0.504664)
    (60,0.520108)
    (70,0.521303)
    (80,0.511639)
    (90,0.514130)
};\addlegendentry{PyLaia}
    \addplot[semithick, mediumpurple152142213, mark=*, mark size=2, mark options={solid}]  coordinates {
    (0,0.966)
    (10,0.961)
    (20,0.960)
    (30,0.955)
    (40,0.949)
    (50,0.949)
    (60,0.950)
    (70,0.948)
    (80,0.952)
    (90,0.980)
};\addlegendentry{Retsinas}

    \end{groupplot}
\node[rotate=90] at ($(group c1r1.west) + (-1.2cm,0)$) {Subset 1};

\node[rotate=90] at ($(group c1r2.west) + (-1.2cm,0)$) {Subset 2};
\node at ($(current bounding box.south)+(0,-0.7cm)$) {
  \pgfplotslegendfromname{mylegend}
};
\end{tikzpicture}
    \caption{Finetuning the models on the aligned training set for different thresholds $\gamma_6$. (a) Normalized CER/WER improvements for the models. The best $\gamma_6$ is a trade off between training data used and confidence (30 -- 60\%), while PyLaia shows the largest improvement.} 
    \label{fig:confidence_eval} 
\end{figure*}

\subsection{Computational Analysis}
We analyze the computational differences regarding the alignment between the models in Table~\ref{tab:compression}. First, the number of time steps per algorithm varies: HTRFlor uses 256 steps per line, while Retsinas et al. use 128. However, they propose inserting whitespace at the beginning and end of each line during training (removed at inference), effectively resulting in 126 steps per line. PyLaia applies adaptive pooling, with an average of 214 steps per line.

Interestingly, we observe that $\varepsilon$-compression reduces the effective sequence length per letter more efficiently for HTRFlor and PyLaia, a result of the sparsity of the CTC probabilities ("peaks") and the relative difference of the steps per line. Overall, we have an average run-time for an alignment of a letter (consisting of 34.6 lines on average) of 2.28 - 3.54s, depending on the algorithm. We conclude that the CTC alignment algorithm can even be applied to longer sequences (or \ac{HTR} models with a larger number of steps) by using the proposed $\varepsilon$-compression. 

\subsection{Self-Training}
After the text alignment, we run the algorithm on the training set of the Bullinger dataset and create subsets to finetune our \ac{HTR} models. Firstly, we evaluate the influence of the confidence threshold. The relative improvements on the baseline \ac{CER} and \ac{WER} metrics are shown in Figure~\ref{fig:confidence_eval}. We notice that the improvement for PyLaia is the largest (which can particularly be observed when evaluating the \ac{CER}$_6$, where the performance improves by 50\%), and the improvement is more pronounced for Subset 1. In general, all models benefit from thresholding, but the gains vary depending on the model and metric. PyLaia exhibits the strongest sensitivity to confidence filtering, while HTRFlor and Retsinas show more moderate but consistent trends.

We observe that the optimal threshold is a trade-off between data quantity and quality, as the highest improvements are obtained when considering samples with a confidence higher than 30--60\%, depending on the approach and subset. This is also in line with the findings by Jungo et al.~\cite{Jungo2023}. When increasing the confidence threshold, the number of training lines decreases significantly (see Figure~\ref{fig:dataset_confidence} in the supplementary), which can negatively affect the finetuning process if too few samples are considered for training. On the other hand, lower thresholds allow for more but also faulty data (see Figure~\ref{fig:accuracy-per-confidence}).

\begin{figure}[hbp]
  \begin{subfigure}[]{\linewidth}
      \centering \scriptsize
      \includegraphics[width=\linewidth]{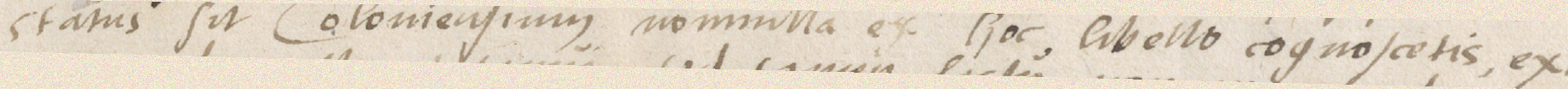} \\
      \textbf{GT} status sit Coloniensium nonnulla ex hoc libello cognoscetis ex \\
      \textbf{FT} \textcolor{green}{status} sit Coloniensium nonnila ex hoc libello \textcolor{green}{cognoscetis ex} \\
      \textbf{BL} sit \textcolor{red}{Coloviensium} nomnia ex hoc libello \textcolor{red}{cognoscetisex}
    \end{subfigure}\\[0.05em]
    

    \begin{subfigure}{\linewidth}
      \centering\scriptsize
      \includegraphics[width=\linewidth]{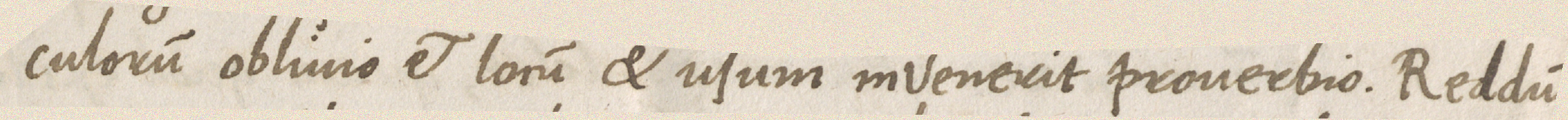} \\
      \textbf{GT} culorum obliuio et lorum \& usum invenerit prouerbio. Redduo \\
      \textbf{FT} \textcolor{green}{clorum} oblivio et locum et usum invenerit proverbio. Reddum \\
      \textbf{BL} oblivio et locum et usum invenerit proverbio. Reddum
    \end{subfigure} \\[0.05em]
    

\begin{subfigure}[]{\linewidth}
  \centering\scriptsize
  \includegraphics[width=\linewidth]{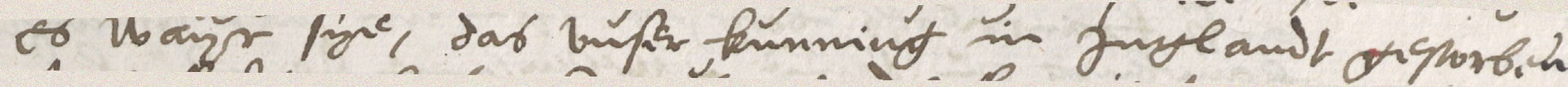} \\
  \textbf{GT} Es wayr sye, das vnser kunning in Inglandt gestorben \\
  \textbf{FT} es waist sye, das unser kunning in \textcolor{red}{iuglandt gestothen} \\
  \textbf{BL} es wait sye, das unser kunning in \textcolor{green}{Inglandt gestorben}
\end{subfigure} \\[0.05em]


\caption{Qualitative comparison of the finetuned (FT) and the baseline  (BL) PyLaia~\cite{PyLaia} model. Red and green words highlight differences that are only correct in one of the predictions.}\label{fig:qual_samples}
\end{figure}

\begin{table}[hbp]
    \centering\small
\begin{tabular}{lcccc}
\toprule
\multirow{2}{*}{} & \multicolumn{2}{c}{Subset 1} & \multicolumn{2}{c}{Subset 2} \\
                 & \ac{CER} & \ac{WER} & \ac{CER} & \ac{WER} \\ 
\midrule 
HTRFlor~\cite{deSousaNeto2020} & 14.8 \crel{0.9} & 51.2 \crel{1.2} & 22.2 \crel{0.6} & 64.4 \crel{0.5} \\
PyLaia~\cite{PyLaia} & \textbf{\phantom{0}6.5} \crel{1.1} & \textbf{25.4} \crel{3.5} & \textbf{17.9} \crel{0.7} & \textbf{52.2} \crel{1.8} \\
Retsinas~\cite{Retsinas2022} & 12.2 \crel{0.8} & 38.8 \crel{1.3} & 21.3 \crel{0.4} & 58.2 \crel{0.4} \\
\bottomrule
\end{tabular}
    \caption{Absolute finetuning improvements for a threshold of 50\%.}
    \label{tab:absolute_htr_improvements}
\end{table}

Table~\ref{tab:absolute_htr_improvements} presents the absolute \ac{CER} and \ac{WER} scores for each model when finetuned using a confidence threshold of 50\%. Consistent with the trends observed in Figure~\ref{fig:confidence_eval}, PyLaia achieves the best overall performance across both subsets and metrics (with an improvement of 1.1\% on the \ac{CER} on Subset 1). However, our algorithm and finetuning strategy improves the results of all three models. We also show qualitative results in Figure~\ref{fig:qual_samples}, where we observe that the finetuned model performs better at the beginning and the end of the line, while the baseline tends to drop the first/last word of the line.

 \paragraph{Line-level Accuracy} Table~\ref{tab:finetuning-line-level-accuracy} reports the line-level alignment accuracy after finetuning the \ac{HTR} models. Overall, the improvements in transcription quality lead to better alignment accuracy for most models. HTRFlor shows the highest post-finetuning accuracy on both subsets, with a gain of 3.4\% on Subset 1. Retsinas et al. also improves across both subsets, though to a lesser extent. Interestingly, PyLaia shows a slight drop in accuracy on Subset 1, likely due to a saturation effect, as its pre-finetuning performance on Subset 1 is already very high. This suggests that for stronger models, further finetuning with filtered data may not always translate into better alignment, especially when initial transcriptions are already of high quality in terms of \ac{CER}. With the improved alignment accuracy, the models can then be finetuned on the enhanced training data again.

\begin{table}[hbp]
    \centering \small
\begin{tabular}{lcc}
\toprule
\multirow{2}{*}{} & Subset 1 & Subset 2 \\

\midrule 
HTRFlor~\cite{deSousaNeto2020} & \textbf{89.5} \mrel{3.4} & \textbf{81.0} \mrel{1.3}\\
PyLaia~\cite{PyLaia} & 85.6 \prel{0.5} & 78.9 \mrel{1.6} \\
Retsinas et al.~\cite{Retsinas2022} & 82.9 \mrel{1.4} & 73.5 \mrel{0.6} \\
\bottomrule
\end{tabular}
    \caption{Line-level accuracy of the alignment after finetuning the \ac{HTR} models.}
    \label{tab:finetuning-line-level-accuracy} 
\end{table}

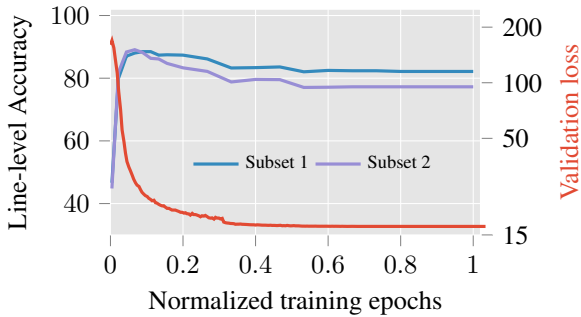
\begin{figure}[htbp]
        \centering
        \begin{tikzpicture}
\definecolor{gainsboro229}{RGB}{229,229,229}
\definecolor{steelblue52138189}{RGB}{52,138,189}
\definecolor{mediumpurple152142213}{RGB}{152,142,213}
\definecolor{lightgray204}{RGB}{204,204,204}
\definecolor{chocolate2267451}{RGB}{226,74,51}

\begin{axis}[
axis background/.style={fill=gainsboro229},
axis line style={white},
    xlabel={Normalized training epochs},
    ylabel={Line-level Accuracy},
    legend style={at={(0.55,0.4)}, anchor=north, fill=none, draw=none, legend columns=-1},
legend cell align={left},
legend style={fill opacity=0, draw opacity=1, text opacity=1, font=\scriptsize},
tick align=outside,
tick pos=left,
ymax=102,ymin=30,
xmin=0.0,xmax=1.02,
x grid style={white},
y grid style={white},
xmajorgrids,
ymajorgrids,
width=0.78\columnwidth,
    height=0.55\columnwidth
]

\addplot[
    color=steelblue52138189, 
    very thick
] coordinates {
(0.004, 46.639)
(0.022, 79.933)
(0.044, 87.029)
(0.067, 88.027)
(0.089, 88.47)
(0.111, 88.47)
(0.133, 87.361)
(0.156, 87.472)
(0.2, 87.361)
(0.267, 86.142)
(0.333, 83.259)
(0.4, 83.37)
(0.467, 83.592)
(0.533, 82.04)
(0.6, 82.483)
(0.667, 82.373)
(0.733, 82.373)
(0.8, 82.151)
(0.867, 82.151)
(0.933, 82.151)
(1.0, 82.151)
};
\addlegendentry{Subset 1}

\addplot[
    color=mediumpurple152142213,
    very thick
] coordinates {
(0.004, 44.818)
(0.022, 81.83)
(0.044, 88.358)
(0.067, 89.098)
(0.089, 88.156)
(0.111, 86.339)
(0.133, 86.137)
(0.156, 84.724)
(0.2, 83.311)
(0.267, 82.167)
(0.333, 78.802)
(0.4, 79.61)
(0.467, 79.542)
(0.533, 77.052)
(0.6, 77.12)
(0.667, 77.254)
(0.733, 77.254)
(0.8, 77.254)
(0.867, 77.254)
(0.933, 77.254)
(1.0, 77.254)
};
\addlegendentry{Subset 2}
\end{axis}

\begin{axis}[
ymode=log,
axis y line*=right,
axis x line=none,
axis line style={white},
x grid style={white},
xmin=0.00, xmax=1,
xtick pos=left,
xtick style={color=white},
y grid style={white},
yticklabel style={anchor=west}, 
ylabel=\textcolor{chocolate2267451}{Validation loss},
ymin=15, ymax=250,
xmin=0.0,xmax=1.02,
ytick={15,50,100,200},
yticklabels={15,50,100,200},
ytick pos=right,
yticklabel style={anchor=west},
width=0.8\columnwidth,
    height=0.55\columnwidth
]
\addplot[
    color=chocolate2267451,
    very thick
] coordinates {
    (0.000, 160.61016845703125)
    (0.004, 169.2061309814453)
    (0.009, 154.67593383789062)
    (0.013, 134.17617797851562)
    (0.018, 113.09540557861328)
    (0.022, 89.60102844238281)
    (0.027, 71.51268768310547)
    (0.031, 55.55810546875)
    (0.036, 47.98185348510742)
    (0.040, 41.66398239135742)
    (0.044, 37.44731521606445)
    (0.049, 34.979427337646484)
    (0.053, 33.47759246826172)
    (0.058, 31.5648193359375)
    (0.062, 30.04094123840332)
    (0.067, 28.675073623657227)
    (0.071, 27.89116096496582)
    (0.076, 27.454195022583008)
    (0.080, 26.421316146850586)
    (0.084, 25.53803062438965)
    (0.089, 24.922645568847656)
    (0.093, 24.49225616455078)
    (0.098, 24.203624725341797)
    (0.102, 23.68783950805664)
    (0.107, 23.336315155029297)
    (0.111, 22.972291946411133)
    (0.116, 23.10926055908203)
    (0.120, 22.413410186767578)
    (0.124, 22.290868759155273)
    (0.129, 21.943742752075195)
    (0.133, 21.729951858520508)
    (0.138, 21.538005828857422)
    (0.142, 21.57184600830078)
    (0.147, 21.157777786254883)
    (0.151, 20.886268615722656)
    (0.156, 20.88232421875)
    (0.160, 20.656150817871094)
    (0.164, 20.4564208984375)
    (0.169, 20.446462631225586)
    (0.173, 20.447463989257812)
    (0.178, 20.252790451049805)
    (0.182, 20.109542846679688)
    (0.187, 19.88395118713379)
    (0.191, 19.886632919311523)
    (0.196, 19.689115524291992)
    (0.200, 19.776161193847656)
    (0.204, 19.541852951049805)
    (0.209, 19.492979049682617)
    (0.213, 19.20765495300293)
    (0.218, 19.624420166015625)
    (0.222, 19.33527946472168)
    (0.227, 19.322681427001953)
    (0.231, 19.205060958862305)
    (0.236, 19.026315689086914)
    (0.240, 18.81238555908203)
    (0.244, 19.04801368713379)
    (0.249, 18.964553833007812)
    (0.253, 19.125221252441406)
    (0.258, 18.837831497192383)
    (0.262, 18.576414108276367)
    (0.267, 18.48927879333496)
    (0.271, 18.554603576660156)
    (0.276, 18.27287483215332)
    (0.280, 18.670995712280273)
    (0.284, 18.505632400512695)
    (0.289, 18.509613037109375)
    (0.293, 18.304473876953125)
    (0.298, 18.437490463256836)
    (0.302, 17.60458755493164)
    (0.307, 17.456314086914062)
    (0.311, 17.405372619628906)
    (0.316, 17.36846351623535)
    (0.320, 17.299922943115234)
    (0.324, 17.286958694458008)
    (0.329, 17.253395080566406)
    (0.333, 17.128440856933594)
    (0.338, 17.152389526367188)
    (0.342, 17.152267456054688)
    (0.347, 17.18138313293457)
    (0.351, 17.127059936523438)
    (0.356, 17.113433837890625)
    (0.360, 17.09469223022461)
    (0.364, 17.07729148864746)
    (0.369, 17.047874450683594)
    (0.373, 17.027807235717773)
    (0.378, 17.014434814453125)
    (0.382, 17.0168514251709)
    (0.387, 16.96181297302246)
    (0.391, 17.013769149780273)
    (0.396, 16.99267578125)
    (0.400, 16.968326568603516)
    (0.404, 16.938827514648438)
    (0.409, 16.935617446899414)
    (0.413, 16.93667984008789)
    (0.418, 16.912097930908203)
    (0.422, 16.9009952545166)
    (0.427, 16.92751121520996)
    (0.431, 16.88353157043457)
    (0.436, 16.892108917236328)
    (0.440, 16.876970291137695)
    (0.444, 16.85537338256836)
    (0.449, 16.88136863708496)
    (0.453, 16.912702560424805)
    (0.458, 16.818187713623047)
    (0.462, 16.842111587524414)
    (0.467, 16.813289642333984)
    (0.471, 16.887691497802734)
    (0.476, 16.827682495117188)
    (0.480, 16.794570922851562)
    (0.484, 16.80710220336914)
    (0.489, 16.843029022216797)
    (0.493, 16.79958152770996)
    (0.498, 16.795879364013672)
    (0.502, 16.793718338012695)
    (0.507, 16.758533477783203)
    (0.511, 16.73826789855957)
    (0.516, 16.73640251159668)
    (0.520, 16.72787857055664)
    (0.524, 16.741722106933594)
    (0.529, 16.720876693725586)
    (0.533, 16.727752685546875)
    (0.538, 16.725563049316406)
    (0.542, 16.726224899291992)
    (0.547, 16.734285354614258)
    (0.551, 16.716711044311523)
    (0.556, 16.713056564331055)
    (0.560, 16.711023330688477)
    (0.564, 16.714920043945312)
    (0.569, 16.712799072265625)
    (0.573, 16.71583366394043)
    (0.578, 16.71500587463379)
    (0.582, 16.706016540527344)
    (0.587, 16.702638626098633)
    (0.591, 16.717588424682617)
    (0.596, 16.699914932250977)
    (0.600, 16.709732055664062)
    (0.604, 16.70553207397461)
    (0.609, 16.69999122619629)
    (0.613, 16.687480926513672)
    (0.618, 16.704307556152344)
    (0.622, 16.69162368774414)
    (0.627, 16.688322067260742)
    (0.631, 16.67901039123535)
    (0.636, 16.683900833129883)
    (0.640, 16.688907623291016)
    (0.644, 16.69458770751953)
    (0.649, 16.69038200378418)
    (0.653, 16.685699462890625)
    (0.658, 16.679256439208984)
    (0.662, 16.679527282714844)
    (0.667, 16.678447723388672)
    (0.671, 16.6806640625)
    (0.676, 16.68183708190918)
    (0.680, 16.681598663330078)
    (0.684, 16.681087493896484)
    (0.689, 16.680622100830078)
    (0.693, 16.68025016784668)
    (0.698, 16.680130004882812)
    (0.702, 16.680009841918945)
    (0.707, 16.68007469177246)
    (0.711, 16.679969787597656)
    (0.716, 16.679872512817383)
    (0.720, 16.679780960083008)
    (0.724, 16.679733276367188)
    (0.729, 16.679622650146484)
    (0.733, 16.67955780029297)
    (0.738, 16.679378509521484)
    (0.742, 16.67929458618164)
    (0.747, 16.679237365722656)
    (0.751, 16.679237365722656)
    (0.756, 16.679153442382812)
    (0.760, 16.678991317749023)
    (0.764, 16.67890739440918)
    (0.769, 16.67889976501465)
    (0.773, 16.67885971069336)
    (0.778, 16.67866325378418)
    (0.782, 16.678577423095703)
    (0.787, 16.678537368774414)
    (0.791, 16.678462982177734)
    (0.796, 16.67835807800293)
    (0.800, 16.67826271057129)
    (0.804, 16.67812728881836)
    (0.809, 16.6781005859375)
    (0.813, 16.67799186706543)
    (0.818, 16.678001403808594)
    (0.822, 16.677818298339844)
    (0.827, 16.677751541137695)
    (0.831, 16.677648544311523)
    (0.836, 16.677614212036133)
    (0.840, 16.67744255065918)
    (0.844, 16.677383422851562)
    (0.849, 16.6772403717041)
    (0.853, 16.677194595336914)
    (0.858, 16.67716407775879)
    (0.862, 16.6771240234375)
    (0.867, 16.676984786987305)
    (0.871, 16.676939010620117)
    (0.876, 16.676929473876953)
    (0.880, 16.67679786682129)
    (0.884, 16.676694869995117)
    (0.889, 16.67655372619629)
    (0.893, 16.676448822021484)
    (0.898, 16.67639923095703)
    (0.902, 16.676345825195312)
    (0.907, 16.676347732543945)
    (0.911, 16.676279067993164)
    (0.916, 16.676132202148438)
    (0.920, 16.6761531829834)
    (0.924, 16.676054000854492)
    (0.929, 16.6760311126709)
    (0.933, 16.67595672607422)
    (0.938, 16.67591667175293)
    (0.942, 16.675809860229492)
    (0.947, 16.67570686340332)
    (0.951, 16.675643920898438)
    (0.956, 16.675580978393555)
    (0.960, 16.67542266845703)
    (0.964, 16.675325393676758)
    (0.969, 16.67531394958496)
    (0.973, 16.6751766204834)
    (0.978, 16.67515754699707)
    (0.982, 16.675039291381836)
    (0.987, 16.67499542236328)
    (0.991, 16.6749210357666)
    (0.996, 16.674806594848633)
    (1.000, 16.674540787935257)
};
\end{axis}
\end{tikzpicture}
        \caption{Line-level accuracy during training}
        \label{fig:weakly-trained-model}
\end{figure}

\subsection{Weakly Trained Model}

Next, we examine the influence of the training steps on the alignment accuracy to check whether a weakly trained model has an improved benefit from the alignment since it has a minor training on the faulty data. The results are shown in Figure~\ref{fig:weakly-trained-model}. The model is trained for 225 epochs, and the probabilities of the weakly trained model do result in a better line-level accuracy -- 88.5\% vs. 86.1\% on Subset 1 and 89.1\% vs. 77.3\% on Subset 2, both reached after 15 epochs of training. We also find that the line-level accuracy does converge after approximately half of the training time in our case. This result highlights that the self-training is more beneficial when the model is trained for a reduced amount of optimization steps, making our approach more efficient with respect to time.

\begin{figure}[htbp]
        \centering
        \begin{subfigure}[b]{0.48\linewidth}
            \centering
            \definecolor{myblue}{RGB}{31,119,180}
\definecolor{gainsboro229}{RGB}{229,229,229}

\begin{tikzpicture}
\begin{axis}[
    ybar,
    axis background/.style={fill=gainsboro229},
    ymax=30.0,ymin=0,
    axis line style={white},
    title={Subset 1},
every axis title/.style={below right,anchor=south,at={(0.5,1)},fill=white},
    width=1.2\columnwidth,
    height=\columnwidth,
    bar width=12pt,   
    enlarge x limits = 0.2,
    every node near coord/.append style={font=\scriptsize},
    symbolic x coords={1.3M,4.9M,6.4M},
    xtick=data,
    nodes near coords,
    nodes near coords align={vertical},
    y axis line style = {draw=none},
    ytick style = {draw=none},
    yticklabels = {},
    xtick style={draw=none}
    ]
\addplot[fill=myblue!30] coordinates {(1.3M,23.3) (4.9M,18.3) (6.4M,7.6)};
\addplot[fill=myblue!80] coordinates {(1.3M,21.6) (4.9M,13.9) (6.4M,6.5)};
\end{axis}
\end{tikzpicture}
        \end{subfigure}
        \begin{subfigure}[b]{0.48\linewidth}
            \centering
            \definecolor{myblue}{RGB}{31,119,180}
\definecolor{gainsboro229}{RGB}{229,229,229}

\begin{tikzpicture}
\begin{axis}[
    ybar,
    axis background/.style={fill=gainsboro229},
    ymax=30.0,ymin=0,
    axis line style={white},
    title={Subset 2},
every axis title/.style={below right,anchor=south,at={(0.5,1)},fill=white},
    width=1.2\columnwidth,
    height=\columnwidth,
    enlarge x limits = 0.2,
    bar width=12pt,                     
    symbolic x coords={1.3M,4.9M,6.4M},
    xtick=data,
    nodes near coords,
    nodes near coords align={vertical},
    every node near coord/.append style={font=\scriptsize},
    y axis line style = {draw=none},
    ytick style = {draw=none},
    yticklabels = {},
    xtick style={draw=none}
    ]
\addplot[fill=myblue!30] coordinates {(1.3M,24.1) (4.9M,23.5) (6.4M,18.6)};
\addplot[fill=myblue!80] coordinates {(1.3M,23.8) (4.9M,21.6) (6.4M,17.9)};
\end{axis}
\end{tikzpicture}
        \end{subfigure}
        \definecolor{myblue}{RGB}{31,119,180}
        \caption{Influence of the model size \colorbox{myblue!30}{before} and \colorbox{myblue!80}{after} finetuning on the \ac{CER}.}
        \label{fig:abl_modelsize}
\end{figure}
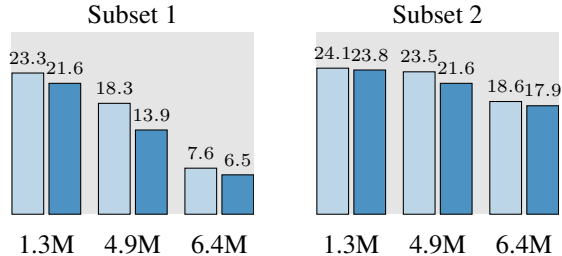

\subsection{Model size}

We evaluate the impact of model size by simplifying the PyLaia architecture; details of the reduced models are in the supplementary. The small model uses a 2-layer RNN with 128 units, while the mid and large models use 3 layers with 256 units. As shown in Figure~\ref{fig:abl_modelsize}, larger models lead to lower CERs both before and after fine-tuning. On Subset 1, the CER drops from 23.3\% to 7.6\% as model size increases from 1.3M to 6.4M parameters (6.5\% after fine-tuning). A similar trend is observed on Subset 2, where CER decreases from 24.1\% to 18.6\% (17.9\% after fine-tuning).


\section{Conclusion}\label{sec:conclusion}
This work introduced a self-training method based on a CTC alignment algorithm to improve annotation quality within the Bullinger dataset. We showed improved performance of three different \ac{HTR} approaches by finetuning on aligned subsets of the initial training data. Our algorithm is efficient in terms of computational complexity, even for longer sequences. Additionally, results indicated that weakly trained models are more suitable for self-training in terms of line-level accuracy, making an iterative process feasible. Lastly, we provided 100 annotated and manually corrected pages of the Bullinger dataset for further research.

For future work, this paper is the first step toward acquiring annotations close to the original visual text. We plan to extend the algorithm to skip words, enabling detection of missing words or abbreviations, common in expert annotations of historical manuscripts. This extension would help obtain character-accurate annotations of the Bullinger data collection to further improve \ac{HTR} performance.

\subsection*{Acknowledgements}
\small This work has been supported by the Hasler Foundation, Switzerland. We thank the contributors of the Bullinger Digital project, in particular Tobias Hodel, Anna Janka, Raphael Müller, Peter Rechsteiner, Patricia Scheurer, David Selim Schoch, Raphael Schwitter, Christian Sieber, Phillip Ströbel, Martin Volk, and Jonas Widmer for their contributions to the dataset and ground truth creation.

\small
\bibliographystyle{ieeenat_fullname}
\bibliography{main}

\newpage\newpage
\phantom{\lipsum[30]}
\pagebreak

\section{Model architectures}
\begin{figure*}[h]
    \centering
    \begin{subfigure}[t]{0.32\textwidth}
        \centering
        \includegraphics[width=0.85\linewidth]{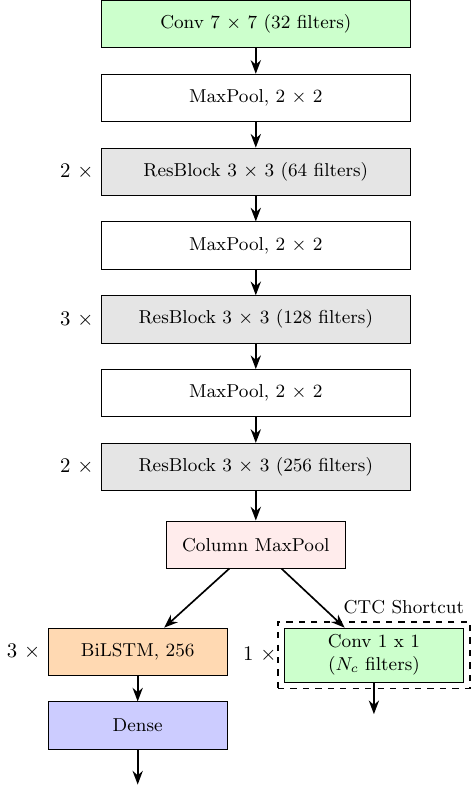}
        \caption{Retsinas et al.~\cite{Retsinas2022}}
        \label{fig:retsinas}
    \end{subfigure}
    \hfill
    \begin{subfigure}[t]{0.32\textwidth}
        \centering
        \includegraphics[width=0.81\linewidth]{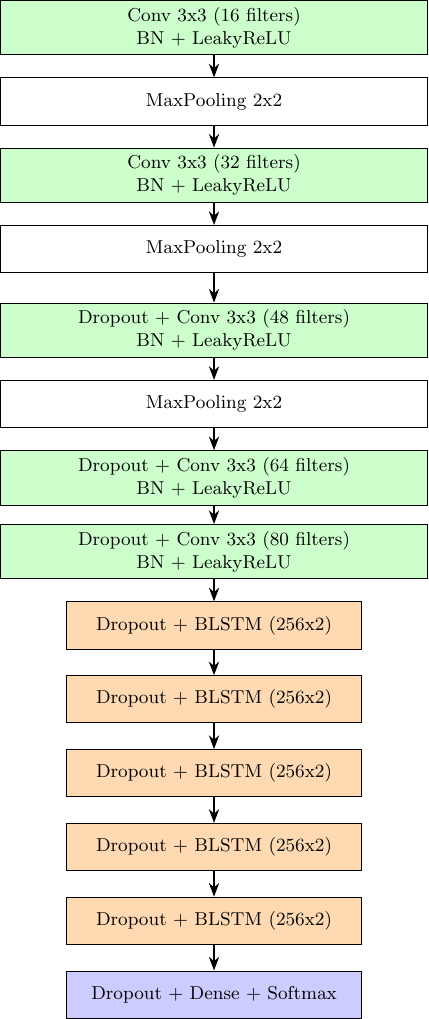}
        \caption{PyLaia~\cite{PyLaia}}
        \label{fig:pylaia}
    \end{subfigure}
    \hfill
    \begin{subfigure}[t]{0.32\textwidth}
        \centering
        \includegraphics[width=0.95\linewidth]{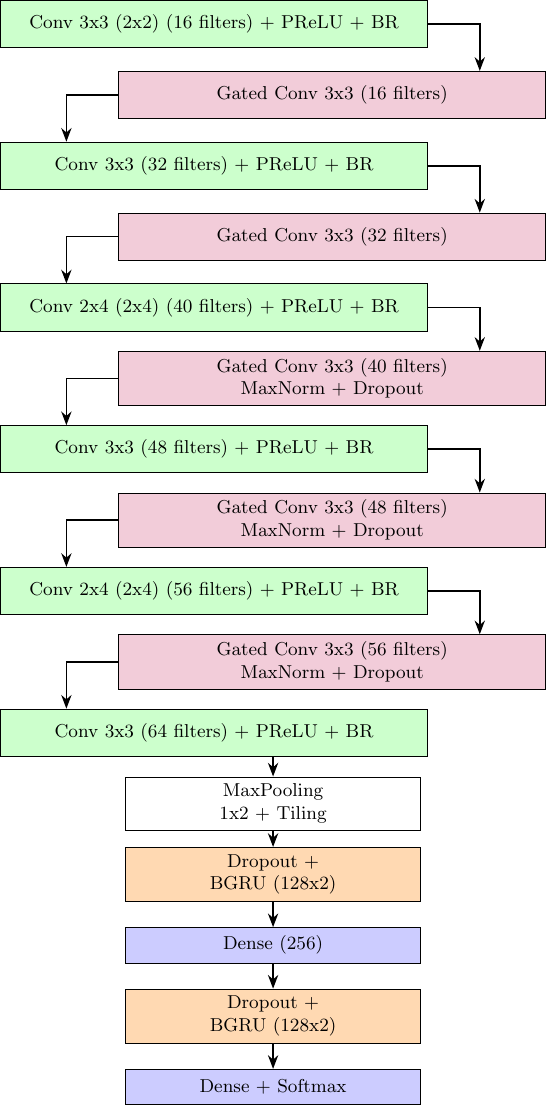}
        \caption{HTRFlor~\cite{deSousaNeto2020}}
        \label{fig:htrflor}
    \end{subfigure}
    \caption{Comparison of the three architectural designs used for text recognition.}
    \label{fig:architecture-comparison}
\end{figure*}

Figure~\ref{fig:architecture-comparison} presents the default architectural configurations of the three \ac{HTR} systems evaluated in this work: Retsinas et al.~\cite{Retsinas2022}, PyLaia~\cite{PyLaia}, and HTRFlor~\cite{deSousaNeto2020}. The model proposed by Recsinas et al. (Figure~\ref{fig:architecture-comparison}a) employs a deep residual convolutional neural network with multiple ResBlocks and intermediate max pooling layers, followed by a column-wise max pooling operation. Sequence modeling is performed by a single BiLSTM layer with 256 hidden units, followed by a dense output layer. Additionally, the architecture incorporates a CTC shortcut path consisting of a $1 \times 1$ convolution for intermediate supervision. PyLaia (Figure~\ref{fig:architecture-comparison}b) adopts a relatively shallow CNN feature extractor composed of five convolutional layers with batch normalization and LeakyReLU activations. This is followed by a stack of five BiLSTM layers, each with 256 hidden units per direction. The final output is produced by a softmax layer applied after a dense transformation.

HTRFlor (Figure~\ref{fig:architecture-comparison}c) is based on gated convolutions and uses PReLU activations combined with Batch Renormalization and MaxNorm regularization. The encoder consists of six gated convolutional blocks with increasing filter sizes, interleaved with dropout and normalization layers. Max pooling and tiling are applied before the recurrent decoder, which comprises two stacked BGU layers, each with 128 hidden units. The final dense layer maps the outputs to character probabilities, followed by a softmax operation. In our experiments, we use the unmodified, default versions of all three architectures as provided in their respective implementations.

\section{PyLaia model architectures}

In Table~\ref{tab:model_comparison}, the detailed architectures of the PyLaia model for the ablation study on the model size are reported. Large corresponds to the default architecture and is also the default provided in the PyLaia Framework. The main difference between the mid and large model is the reduced size of the CNN backbone (3 vs. 4), while the small and mid model differ mainly in terms of the reduced size of the RNN.

\begin{table}[h]
\centering \small
\begin{tabular}{lccc}
\toprule
\textbf{Parameter} & \textbf{Small} & \textbf{Mid} & \textbf{Large} \\
\midrule
\textbf{CNN} & ~ & ~ & ~ \\ 
Features       & [16, 24, 48]     & [16, 24, 36]     & [16, 16, 32, 32] \\
Kernel Size    & [3, 3, 3]        & [3, 3, 3]        & [3, 3, 3, 3] \\
Stride         & [1, 1, 1]        & [1, 1, 1]        & [1, 1, 1, 1] \\
Dilation       & [1, 1, 1]        & [1, 1, 1]        & [1, 1, 1, 1] \\
Pool Size      & [2, 2, 2]        & [2, 2, 2]        & [2, 2, 2, 0] \\
Dropout        & 0 & 0 & 0 \\

Activation     & LeakyReLU    & LeakyReLU   & LeakyReLU \\ \midrule
\textbf{RNN} & ~ & ~ & ~ \\ 
Layers         & 2                & 3                & 3 \\
Units          & 128              & 256              & 256 \\
Type           & LSTM             & LSTM             & LSTM \\
Dropout        & 0.5              & 0.5              & 0.5 \\ \midrule
\textbf{Linear} & ~ & ~ & ~ \\

Dropout     & 0.5              & 0.5              & 0.5 \\
\bottomrule
\end{tabular}
\caption{Comparison of Small (1.3M), Mid (4.9M), and Large (6.4M) PyLaia model configurations. (T = True, F = False)}
\label{tab:model_comparison}
\end{table}

\subsection{Subset size for finetuning}
\begin{figure*}[htbp]
    \centering
\begin{tikzpicture}

\definecolor{chocolate2267451}{RGB}{226,74,51}
\definecolor{dimgray85}{RGB}{85,85,85}
\definecolor{gainsboro229}{RGB}{229,229,229}
\definecolor{lightgray204}{RGB}{204,204,204}
\definecolor{mediumpurple152142213}{RGB}{152,142,213}
\definecolor{steelblue52138189}{RGB}{52,138,189}

\begin{axis}[
axis background/.style={fill=gainsboro229},
axis line style={white},
width=0.95\textwidth,
height=0.25\textwidth,
legend cell align={left},
legend style={fill opacity=0.8, draw opacity=1, text opacity=1, draw=lightgray204, fill=gainsboro229},
tick align=outside,
tick pos=left,
x grid style={white},
xlabel=\textcolor{black}{Confidence $\gamma_6 > x$},
ylabel=\textcolor{black}{Number of samples},
xmajorgrids,
xmin=-0.6125, xmax=10.1125,
xtick style={color=dimgray85},
xtick={0.25,1.25,2.25,3.25,4.25,5.25,6.25,7.25,8.25,9.25},
xticklabels={0,10,20,30,40,50,60,70,80,90},
y grid style={white},
ymajorgrids,
ymin=0, ymax=117.6087,
ytick style={color=dimgray85},
ytick={0,25,50,75,100},
yticklabels={0,25k,50k, 75k,100k}
]
\draw[draw=black,fill=chocolate2267451,very thin] (axis cs:-0.125,0) rectangle (axis cs:0.125,106.09);
\addlegendimage{ybar,ybar legend,draw=none,fill=chocolate2267451,very thin}
\addlegendentry{HTRFlor}

\draw[draw=black,fill=chocolate2267451,very thin] (axis cs:0.875,0) rectangle (axis cs:1.125,101.9);
\draw[draw=black,fill=chocolate2267451,very thin] (axis cs:1.875,0) rectangle (axis cs:2.125,101.006);
\draw[draw=black,fill=chocolate2267451,very thin] (axis cs:2.875,0) rectangle (axis cs:3.125,98.984);
\draw[draw=black,fill=chocolate2267451,very thin] (axis cs:3.875,0) rectangle (axis cs:4.125,94.991);
\draw[draw=black,fill=chocolate2267451,very thin] (axis cs:4.875,0) rectangle (axis cs:5.125,88.205);
\draw[draw=black,fill=chocolate2267451,very thin] (axis cs:5.875,0) rectangle (axis cs:6.125,77.946);
\draw[draw=black,fill=chocolate2267451,very thin] (axis cs:6.875,0) rectangle (axis cs:7.125,63.103);
\draw[draw=black,fill=chocolate2267451,very thin] (axis cs:7.875,0) rectangle (axis cs:8.125,42.442);
\draw[draw=black,fill=chocolate2267451,very thin] (axis cs:8.875,0) rectangle (axis cs:9.125,17.369);
\draw[draw=black,fill=steelblue52138189,very thin] (axis cs:0.125,0) rectangle (axis cs:0.375,106.917);
\addlegendimage{ybar,ybar legend,draw=none,fill=steelblue52138189,very thin}
\addlegendentry{PyLaia}

\draw[draw=black,fill=steelblue52138189,very thin] (axis cs:1.125,0) rectangle (axis cs:1.375,95.913);
\draw[draw=black,fill=steelblue52138189,very thin] (axis cs:2.125,0) rectangle (axis cs:2.375,92.142);
\draw[draw=black,fill=steelblue52138189,very thin] (axis cs:3.125,0) rectangle (axis cs:3.375,89.694);
\draw[draw=black,fill=steelblue52138189,very thin] (axis cs:4.125,0) rectangle (axis cs:4.375,86.069);
\draw[draw=black,fill=steelblue52138189,very thin] (axis cs:5.125,0) rectangle (axis cs:5.375,79.805);
\draw[draw=black,fill=steelblue52138189,very thin] (axis cs:6.125,0) rectangle (axis cs:6.375,70.158);
\draw[draw=black,fill=steelblue52138189,very thin] (axis cs:7.125,0) rectangle (axis cs:7.375,57.332);
\draw[draw=black,fill=steelblue52138189,very thin] (axis cs:8.125,0) rectangle (axis cs:8.375,40.538);
\draw[draw=black,fill=steelblue52138189,very thin] (axis cs:9.125,0) rectangle (axis cs:9.375,17.86);
\draw[draw=black,fill=mediumpurple152142213,very thin] (axis cs:0.375,0) rectangle (axis cs:0.625,106.741);
\addlegendimage{ybar,ybar legend,draw=none,fill=mediumpurple152142213,very thin}
\addlegendentry{Retsinas}

\draw[draw=black,fill=mediumpurple152142213,very thin] (axis cs:1.375,0) rectangle (axis cs:1.625,95.076);
\draw[draw=black,fill=mediumpurple152142213,very thin] (axis cs:2.375,0) rectangle (axis cs:2.625,92.395);
\draw[draw=black,fill=mediumpurple152142213,very thin] (axis cs:3.375,0) rectangle (axis cs:3.625,86.45);
\draw[draw=black,fill=mediumpurple152142213,very thin] (axis cs:4.375,0) rectangle (axis cs:4.625,77.365);
\draw[draw=black,fill=mediumpurple152142213,very thin] (axis cs:5.375,0) rectangle (axis cs:5.625,66.636);
\draw[draw=black,fill=mediumpurple152142213,very thin] (axis cs:6.375,0) rectangle (axis cs:6.625,54.38);
\draw[draw=black,fill=mediumpurple152142213,very thin] (axis cs:7.375,0) rectangle (axis cs:7.625,40.602);
\draw[draw=black,fill=mediumpurple152142213,very thin] (axis cs:8.375,0) rectangle (axis cs:8.625,24.124);
\draw[draw=black,fill=mediumpurple152142213,very thin] (axis cs:9.375,0) rectangle (axis cs:9.625,7.631);
\end{axis}

\end{tikzpicture}
    \caption{Number of training samples used for finetuning for different thresholds.} 
    \label{fig:dataset_confidence} 
  \end{figure*}
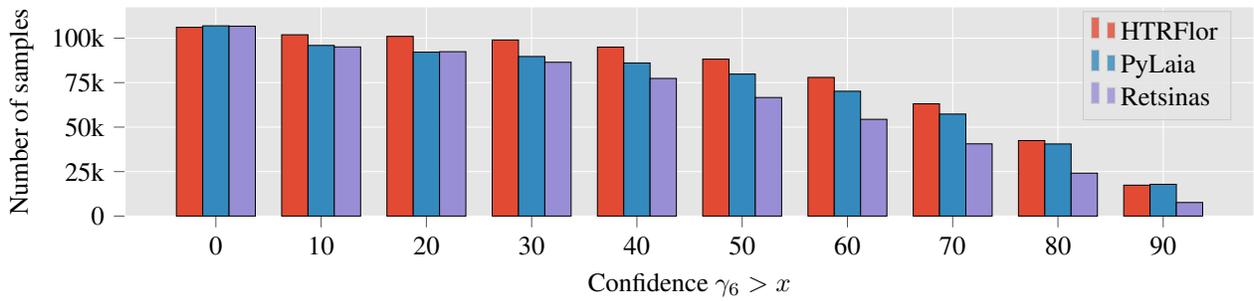

Figure~\ref{fig:dataset_confidence} illustrates the number of training samples retained for finetuning at various confidence thresholds. As the confidence threshold increases from 0 to 90, the number of samples decreases across all three models: HTRFlor, PyLaia, and Retsinas. HTRFlor consistently maintains a higher number of samples compared to the other models at most thresholds, while Retsinas shows the steepest decline. This shows how stricter confidence criteria reduce the training dataset size, which impacts the model finetuning by reducing the amount of faulty data (while also having less data).

\end{document}